\documentclass{article}
\usepackage{iclr2027_conference,times}

\usepackage{amsmath,amsfonts,bm}

\def\eqref#1{equation~\ref{#1}}

\def\1{\bm{1}}

\DeclareMathAlphabet{\mathsfit}{\encodingdefault}{\sfdefault}{m}{sl}
\SetMathAlphabet{\mathsfit}{bold}{\encodingdefault}{\sfdefault}{bx}{n}

\usepackage{hyperref}
\usepackage{url}
\usepackage{graphicx}
\usepackage{booktabs}
\usepackage{multirow}
\usepackage{float}
\usepackage{amsmath}
\usepackage{amssymb}
\usepackage{subcaption}
\usepackage{xcolor}

\graphicspath{{figures/}}

\newcommand{\ours}{\textsc{Ours}}
\newcommand{\method}{\textsc{Manifold4D}}
\newcommand{\bench}{\textsc{DAVIS-Traj}}

\newcommand{\xrender}{x_{\mathrm{render}}}

\newcommand{\xtgt}{x_{\mathrm{tgt}}}

\newcommand{\best}[1]{\textbf{#1}}
\newcommand{\second}[1]{\underline{#1}}
\newcommand{\dg}{$^{\circ}$}

\title{Manifold4D: Denoising on Point Cloud Rendered Manifolds for Video Re-shooting}

\author{
  Yongqi Mao$^{1,2}$, Zijia Dai$^{2,3}$, Zhishuo Liu$^{2,4}$, Wei Xu$^{2}$, Kaiwei Wang$^{1}$, Guotao Meng$^{2}$ \\
  $^{1}$Zhejiang University \quad $^{2}$Manifold Tech \quad $^{3}$ShanghaiTech University \quad $^{4}$University of Cambridge 
}

\newif\ifarxivpreprint
\arxivpreprinttrue
\ifarxivpreprint\iclrfinalcopy\fi

\begin{document}

\maketitle
\ifarxivpreprint\lhead{Preprint}\fi
\begin{figure}[htbp]
  \centering
  \includegraphics[width=0.88\linewidth]{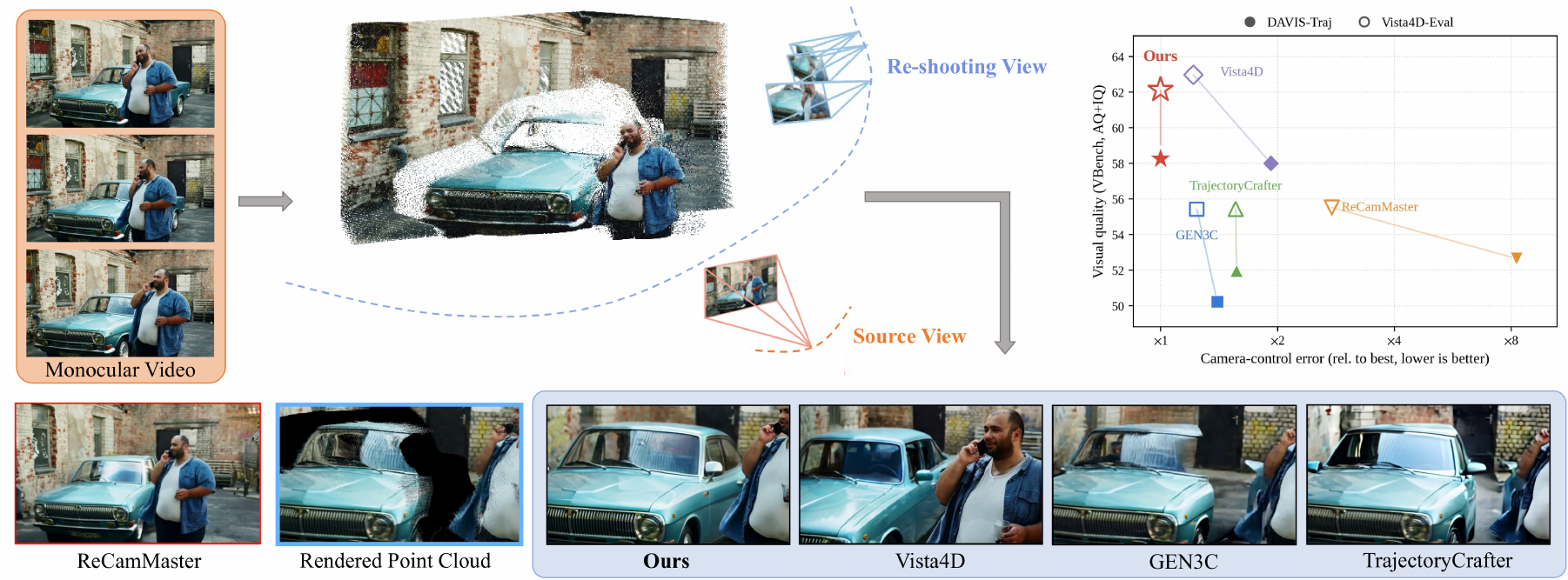}
  \caption{\textbf{Video re-shooting.} Given a monocular video and a target camera trajectory,
  \method{} re-shoots the video along the new trajectory, achieving the best trajectory control
  among existing methods while preserving visual quality.
  The bottom panel shows a re-shooting example.
  GEN3C and TrajectoryCrafter inherit the artifacts of the imperfect render,
  while Vista4D leaves the subject misaligned with the render geometry.
  In the top-right plot, the vertical axis averages the VBench aesthetic and
  imaging quality scores; the horizontal axis denotes the camera-control error
  relative to the best method (see Appendix~\ref{app:details} for details).}
  \label{fig:teaser}
\end{figure}

\begin{abstract}
Video re-shooting is the task of re-rendering a monocular video of a dynamic scene along a user-specified camera trajectory. A dominant paradigm explicitly provides the target geometry by lifting the source video into a 4D point cloud using per-frame depth, and rasterizing this representation along the target trajectory to obtain point-cloud renders as geometric conditions for trajectory-controlled generation.
The point cloud render and source video may provide conflicting visual cues throughout denoising, as both are directly used as conditions. We refer to this conflict as a trust dilemma, which can compromise trajectory control, 3D consistency of dynamic content, and visual quality, particularly under distribution shifts.
We argue that because a point-cloud render is pixel-aligned with the target view, it does not need to be continuously provided as a conditioning signal throughout denoising.
We propose \method{}, which injects the render directly into the initial noise of
flow matching, so that generation
no longer departs from standard Gaussian noise but from a geometry-bearing
noise distribution that we term the point cloud rendered manifold, leaving the
source video as the only visual
condition.
The render is thus used exactly once, and the network is never asked to learn
how to read it; in subsequent denoising steps the model can focus on the source
video.
In our experiments, \method{} attains
the best camera-control accuracy on every metric, lowering rotation error by
$25$--$27\%$ and translation error
by up to $32\%$ over the strongest baseline, while matching it in video
fidelity and attaining the best overall photometric quality and 3D consistency of
dynamic content.
\end{abstract}

\section{Introduction}
\label{sec:intro}

Camera motion is one of the oldest devices of visual storytelling: an
orbiting shot or a well-timed dolly can turn an ordinary take into a
cinematic one.
\emph{Video re-shooting}, camera-controlled video generation from a single
video, brings this freedom to post-production.
Given a monocular video of a dynamic scene and a user-specified camera
trajectory, the goal is to render the same take from new viewpoints:
reconstructing faithfully what the source video observes, synthesizing
plausibly what it does not, and following the requested camera motion
precisely.
It also supplies a multi-view generative prior for monocular 4D
reconstruction and gives world models an explicit handle on viewpoint motion.

To equip a pretrained video diffusion model with such camera-control, the camera
trajectory has to be handed to the network in some form, and existing designs
differ mainly in how explicit that form is.
ReCamMaster~\citep{recammaster} keeps the trajectory implicit: the poses are
compressed into an embedding, and the network is left to infer the implied
geometry, which makes accurate control challenging.
A more explicit line of work first lifts the source video into a point cloud using per-frame depth, then rasterizes it along the target trajectory to obtain rendered frames as geometric guidance, with the source video providing appearance and style guidance.
GEN3C~\citep{ren2025gen3c} and TrajectoryCrafter~\citep{yu2025trajectorycrafter}
concatenate this render with the noisy latent at every denoising step, while the
source video enters only weakly through a first-frame anchor or cross-attention.
The render is often fragmented by sparse point clouds and distorted by depth
errors, so these artifacts propagate into the output.
Vista4D~\citep{vista4d} instead balances the two as token streams of equal
status, keeping visual quality and trajectory control intact, but adds no
constraint enforcing the render geometry, so the subject can drift away from the
target view.
Fig.~\ref{fig:teaser} illustrates both failure modes.

\begin{figure}[t]
  \centering
  \includegraphics[width=\linewidth]{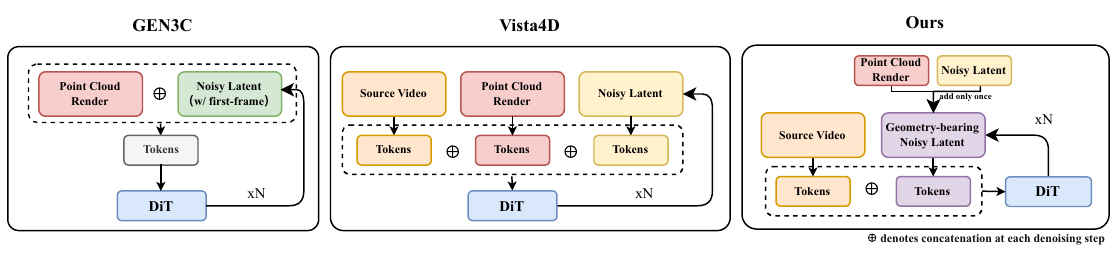}
  \caption{\textbf{How each method consumes the render.} 
  Prior methods concatenate the render at every denoising step, 
  so the network must learn to interpret and utilize it.
  Ours directly adds it into the starting noisy latent,
  requiring no learning.
  This explicit, one-time consumption is the key difference.
  }
  \label{fig:method_differ}
\end{figure}

We refer to this tension as a trust dilemma between the source video
and the point cloud render: which one to believe, and how much. Trusting the render too little weakens trajectory control and 3D consistency, while trusting it too much propagates its artifacts into the output. More fundamentally, this dilemma arises from asking the model to continuously reconcile two visual conditions, a balance that may not generalize beyond the motion magnitudes seen during training.
A natural way to resolve this dilemma is to reduce their interaction during denoising.
Independently and contemporaneously, MoCam~\citep{mocam} addresses this dilemma by assigning the render and source condition to different denoising stages. However, the render remains visible over multiple denoising steps, still requiring the network to learn how to interpret and utilize geometry that is already explicitly given.
Despite depth errors and holes, the point cloud render is already in the target view and pixel-aligned with the video to be produced. We therefore argue that the network need not repeatedly learn how to interpret and utilize this explicit geometric information.
Instead, we add the render to the initial noise, while the source video guides subsequent denoising.
Fig.~\ref{fig:method_differ} contrasts the three designs in how the render enters denoising.

We propose \method{}.
The design philosophy is simple: remove the trust dilemma rather than balance the two signals.
The two signals enter at different points and never compete during denoising.
The render is added into the initial noise a single time, which displaces the
starting distribution off the Gaussian prior and onto the point cloud rendered
manifold.
The source video enters at the token level and remains the only visual condition,
so the network can spend its attention on appearance and detail rather than on
weighing two competing signals, and retains ample room to correct an erroneous render.
This design also reduces training cost, as the network no longer needs to learn how to interpret and utilize the render as a visual condition.
It also infers about $1.5\times$ faster than Vista4D with the same base model
and output format (Appendix~\ref{app:inference}).
We further show experimentally that such an injection does not destroy the diffusion prior: the model adapts to it after fine-tuning.
The simplest entry for the render turns out to be the strongest: trained with
an identical budget, the natural alternatives that keep the render as a
persistent condition fall behind in camera control
(Sec.~\ref{sec:exp:ablation}).

We conduct extensive experiments on multiple benchmarks.
\method{} ranks first on every camera-control metric of both benchmarks while
matching the strongest baseline in visual quality, lowering rotation error by
$25\%$ and $27\%$ over the best baseline on each and translation error by up to
$32\%$, and it achieves clear advantages in trajectory following and 3D consistency of the dynamic subject in our user study.
The advantage widens as the camera motion grows well beyond the training
range, up to a $180$\dg\ change in viewing angle.
We further report a robustness test in which the point cloud render contains
clear errors, and show that our model is still able to correct them from the
source video.
Our contributions are summarized as follows:
\begin{itemize}
  \item We analyze the trust dilemma faced by prior work and offer a new perspective: for pixel-aligned geometric priors, directly incorporating them into the starting state can be more effective than providing them as persistent conditions.
  \item We propose \method{}, a video re-shooting model 
  that injects the point cloud render into the starting point of flow matching 
  so that generation departs from geometry-bearing noise and attains both visual quality and trajectory control.
  \item We conduct extensive experiments showing that \method{} approaches
  or surpasses the pose-estimation accuracy of the point cloud render itself
  (treated as a geometric reference) while preserving visual quality and 3D consistency.
\end{itemize}

\section{Related Work}
\label{sec:related}

\paragraph{Pose-conditioned video generation.}
One line of work encodes the target trajectory into a numerical signal, such as
extrinsic matrices or pixel-wise Pl\"ucker embeddings, and appends it to the
conditioning stream, leaving the network to infer the implied geometry on its
own.
Where no source video is available, this yields camera-controlled text- and
image-to-video
models~\citep{liu2023zero,wang2024motionctrl,he2025cameractrl,bahmani2025vd3d}.
The same recipe carries over to video re-shooting, with the source video
supplied as an additional
reference~\citep{vanhoorick2024gcd,recammaster,zhang2025recapture,camclonemaster,reangle,sv4d}.
In both settings a pose encoding states where the camera goes but not what it
should see, so the mapping from camera parameters to pixels must be learned
entirely from data, which yields weak trajectory control that cannot be previewed
before generation.

\paragraph{Render-conditioned video generation.}
A second line of work makes the geometry explicit: per-frame depth lifts the source
video into a point cloud, which is rasterized along the target trajectory into a
point cloud render. This render is a far more explicit condition than a pose
encoding, since it is pixel-aligned with the target view and states what should
appear at every pixel.
This pipeline was established for static scenes from single or sparse
images~\citep{chan2023generative,wu2024reconfusion,yu2025viewcrafter}, then
carried to monocular video by TrajectoryCrafter~\citep{yu2025trajectorycrafter} and
GEN3C~\citep{ren2025gen3c}.
Both concatenate the render channel-wise with the noisy latent, and differ only in the appearance reference: the whole source video in the former, its first frame alone in the latter.
EX-4D~\citep{ex4d} instead lifts the source scene into a watertight mesh
for extreme viewpoint 4D synthesis, while Vista4D~\citep{vista4d} builds a 4D point
cloud and concatenates render, source video and the noisy latent at the token level.
Since the render and the source video are both explicit visual conditions, they
readily compete.
Independently and contemporaneously with our work, MoCam~\citep{mocam}
identifies this conflict and separates the two signals in time, handing over from render to source video partway through denoising.
Scheduled or static, all methods above ask the network to learn to read a signal that is already given.
We instead inject the render into the initial noise, so that geometry is carried
by the generative state rather than repeatedly supplied as an explicit
conditioning stream, with fine-tuning that matches the resulting distribution.

\paragraph{Signal injection as a starting point.}
SDEdit~\citep{meng2022sdedit} introduces the paradigm of adding noise to an
existing signal and using the result as the starting point for diffusion, rather
than supplying the signal as a condition; the principle generalizes beyond
Gaussian noise~\citep{bansal2023cold} and applies to inpainting~\citep{lugmayr2022repaint}.
Image restoration exploits the same idea: initializing the trajectory from the
degraded observation shortens the generative path and improves
fidelity~\citep{chung2022ccdf,delbracio2023indi,liu2023i2sb}.
Across different models and tasks, these works establish signal injection as an effective starting point; 
we carry this idea into video generation 
and propose \method{}, which injects the point cloud render into the starting
point of flow matching, realizing a new approach to video re-shooting.

\begin{figure}[!t]
  \centering
  \includegraphics[width=\linewidth]{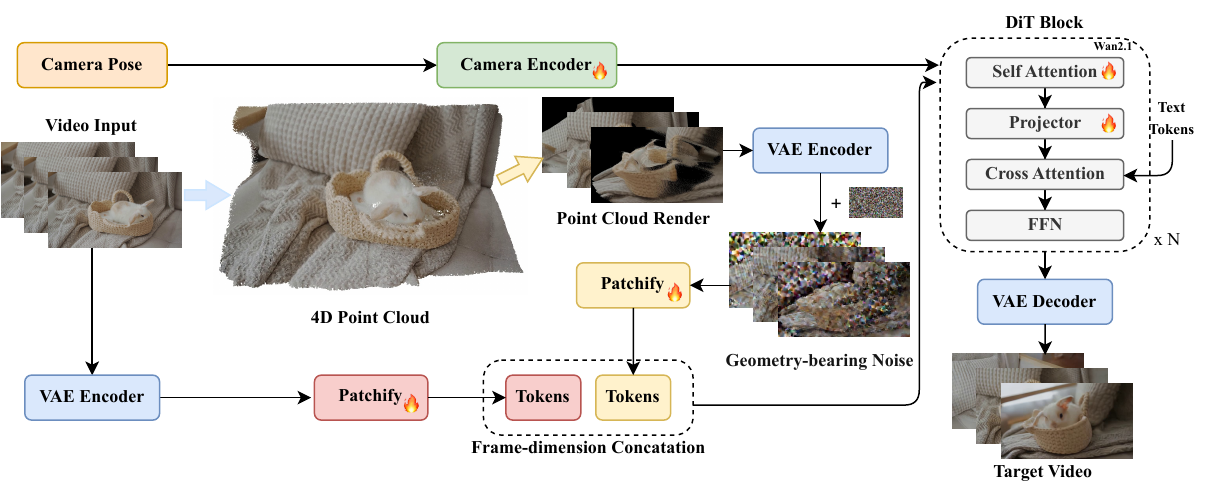}
  \caption{\textbf{Pipeline overview.} Given a monocular video and a target camera trajectory,
   \method{} takes a 4D point cloud obtained from an existing reconstruction and motion decomposition pipeline and renders target-trajectory views from it. The rendered views are encoded into VAE latents, which are injected into the noise and patchified into tokens. The source video is similarly encoded into
   tokens, concatenated with the noise tokens, and fed to the DiT for denoising, together with the
   encoded camera trajectory.}
  \label{fig:pipeline}
\end{figure}

\section{Method}
\label{sec:method}

\subsection{Flow matching for video diffusion}
\label{sec:method:prelim}
We build on Wan2.1-T2V~\citep{wan2025}, a latent video diffusion transformer trained
with flow matching~\citep{lipman2023flow}.
A VAE first compresses the video spatially and temporally, and the velocity
field is learned in this latent space.

Flow matching learns a velocity field that transports a prior distribution to the
data distribution along a straight path.
We follow the convention that $t\!=\!0$ is clean data and $t\!=\!1$ is the prior,
so for a target latent $\xtgt$ and Gaussian noise
$\varepsilon\!\sim\!\mathcal{N}(0,I)$ the training path and its velocity are
\begin{equation}
    x_t = (1-t)\,\xtgt + t\,\varepsilon,
    \qquad
    v^{\star} = \varepsilon - \xtgt ,
    \label{eq:fm}
\end{equation}
and the network $v_\theta$ is trained to regress $v^{\star}$ under
$\mathbb{E}_{t,\xtgt,\varepsilon}\|v_\theta(x_t,t,c) - v^{\star}\|^2$ where $c$ denotes the conditioning input.
For a standard text-to-video model, $c$ is solely the text prompt, while
trajectory-controlled variants place both the source video and the render in
$c$: the trust dilemma analyzed in Sec.~\ref{sec:intro}.
We fine-tune Wan2.1-T2V differently: the render is injected into the starting
point of the path at $t{=}1$, while the source video remains the only visual
condition.

\subsection{Explicit 4D point cloud}
\label{sec:method:pointcloud}
Following the explicit-control paradigm, we first reconstruct a 4D point cloud
and project it along the target trajectory to obtain a rendered video
$\xrender$.
Specifically, we estimate the per-frame depth $D_i$, camera intrinsics $K_i$,
and extrinsics $T_i=[R_i \mid \mathbf{t}_i]$ using VGGT-Omega~\citep{vggtomega},
an extension of VGGT~\citep{vggt},
and back-project every pixel of every source frame into a globally aligned
point cloud.

To enable 4D reconstruction, we further segment the dynamic subject using 
Qwen2.5-VL~\citep{qwen25vl} and SAM3~\citep{sam3} to obtain a motion mask that
separates dynamic points from the static background.
Static points are accumulated over all source frames into one global map
$\mathcal{P}_{\mathrm{static}}$, whereas dynamic points are kept per timestamp,
$\mathcal{P}_{\mathrm{dyn}}^{(i)}$, because their world position changes over
time.
Rendering target frame $j$ therefore rasterizes $\mathcal{P}_{\mathrm{static}}
\cup \mathcal{P}_{\mathrm{dyn}}^{(j)}$ through the target camera
$(K_j, T_j)$, yielding the rendered video $\xrender$.
Rasterization also produces a binary mask indicating whether each pixel receives at least one point. 
We average-pool this mask to the token resolution to obtain the coverage $\alpha\!\in\![0,1]$ for each token, 
which serves as an indicator for distinguishing regions that require generation from those that only need refinement.

\subsection{Geometry-bearing starting point}
\label{conditioning}

The 4D point cloud provides the geometric structure, while the source video
serves as the reference for appearance, style, and fine-grained details.
The two therefore enter the model in different ways: geometry initializes
generation, and appearance guides denoising.

Unlike conventional approaches that concatenate the source video, noise, and
point-cloud rendering as three token streams, we inject the render into the
noise to obtain a geometry-bearing noise representation:
\begin{equation}
    \tilde{x}_1 = \xrender + \sigma\,\varepsilon,
    \qquad
    x_1 = \alpha\,\tilde{x}_1 + (1-\alpha)\,\varepsilon,
    \label{eq:init}
\end{equation}
where \(\varepsilon\!\sim\!\mathcal{N}(0,I)\) is standard Gaussian noise,
\(\tilde{x}_1\) is the starting state of covered tokens, i.e., the
render plus a residual noise of strength \(\sigma\), and
\(\alpha\) the per-token coverage defined above.
Hole tokens therefore start from pure Gaussian noise, fully covered tokens
from \(\xrender+\sigma\,\varepsilon\), and partial coverage interpolates the two.
Training follows Eq.~\eqref{eq:fm} with the Gaussian endpoint replaced by this
geometry-bearing noise, so the regressed velocity becomes
$v^{\star} = x_1 - \xtgt$.
The objective is otherwise unchanged: $v_\theta$ regresses $v^{\star}$ on all
tokens, transporting the point cloud rendered manifold, rather than the
Gaussian prior, to the data distribution.
Because diffusion models lay down geometric structure in their early steps,
the explicit injection lets generation start with this skeleton already in
place, freeing the network's capacity for appearance and detail.

We then concatenate the geometry-bearing noise with the source video at the
token level, $\mathbf{z} = [\,\mathbf{x}_1\,;\,\mathrm{Patchify}(\mathrm{Enc}(S))\,]$,
where $\mathrm{Enc}(S)$ is the VAE-encoded source-video latent.

Because the render is dropped for part of training, we additionally supply the
target trajectory as per-pixel Pl\"ucker embeddings through zero-initialized
linear projections~\citep{recammaster}.
Removing them at inference leaves accuracy and quality essentially unchanged
(Table~\ref{tab:ablation_infer}); we retain them as a training-time safeguard.

Our key insight is that explicit geometry is most effective at the starting
point of the generative path rather than as a persistent condition: since the
render is pixel-aligned with the target view, consuming it there exploits this
alignment directly, and a persistent render condition would also compete with the
source video condition.
One may worry that such an injection breaks the diffusion prior; our
experiments show that the model adapts to this shift through
fine-tuning.

\subsection{Training detail}
\label{sec:method:training}
We build on Wan2.1-T2V-14B and fine-tune patchify layers, self-attention layers, 
camera encoders, and projectors, while freezing all other parameters.
We use AdamW with a constant learning rate of $10^{-5}$ and a global batch size of $8$ for 30K steps, 
at 49 frames and $384\!\times\!672$.
\paragraph{Dataset.}Training mixes five sources (DL3DV~\citep{ling2024dl3dv}, DynPose~\citep{dynpose100k},
OpenVid-HD~\citep{openvid}, MultiCamVideo~\citep{recammaster}, and HuMMan~\citep{cai2022humman};
about $36$K clips), with depth and camera poses from VGGT-Omega for every source
except HuMMan, which ships with calibrated captures; further details in
Appendix~\ref{app:train}.

\section{Experiments}
\label{sec:exp}
\begin{table}[t]
\centering
\caption{\textbf{Camera control accuracy.} \method{} attains the highest accuracy
on every metric, approaching or surpassing the point cloud render itself on
rotation error.
\best{Bold} denotes the best and \second{underline} the second-best result.}
\label{tab:main_pose}
\small
\setlength{\tabcolsep}{3pt}
\renewcommand{\arraystretch}{0.9}
\resizebox{\textwidth}{!}{
\begin{tabular}{lccccccccc}
\toprule
\multirow{2}{*}{Method} & \multicolumn{3}{c}{\bench{} (72 clips)}
 & \multicolumn{3}{c}{Vista4D-Eval (103 clips)}
 & \multicolumn{3}{c}{Vista4D-Eval raw (110 clips)} \\
\cmidrule(lr){2-4}\cmidrule(lr){5-7}\cmidrule(lr){8-10}
 & RotErr\,$\downarrow$ & TransErr\,$\downarrow$ & IntrErr\,$\downarrow$
 & RotErr\,$\downarrow$ & TransErr\,$\downarrow$ & IntrErr\,$\downarrow$
 & RotErr\,$\downarrow$ & TransErr\,$\downarrow$ & IntrErr\,$\downarrow$ \\
\midrule
ReCamMaster       &            14.04 &            0.581 &             2.77 &            11.81 &            0.495 &            10.63 &            13.73 &            0.549 &            11.23 \\
TrajectoryCrafter &             1.56 &            0.274 &             1.14 &             5.40 &            0.462 &             8.40 &             8.24 &            0.518 &             9.14 \\
GEN3C             &    \second{1.45} &   \second{0.236} &    \second{1.13} &    \second{3.77} &            0.441 &             8.74 &    \second{4.31} &            0.460 &             9.33 \\
Vista4D           &             2.56 &            0.240 &             1.47 &             3.89 &   \second{0.403} &    \second{7.66} &             4.84 &   \second{0.432} &    \second{8.10} \\
\midrule
\ours{}           &      \best{1.09} &     \best{0.161} &      \best{1.04} &      \best{2.76} &     \best{0.394} &      \best{7.65} &      \best{4.26} &     \best{0.424} &      \best{8.09} \\
\midrule
Point cloud render & \textit{1.09} & \textit{0.094} & \textit{1.02} & \textit{3.59} & \textit{0.328} & \textit{6.91} & \textit{4.18} & \textit{0.390} & \textit{7.20} \\
\bottomrule
\end{tabular}}
\end{table}

\begin{figure}[t]
  \centering
  \includegraphics[width=\linewidth]{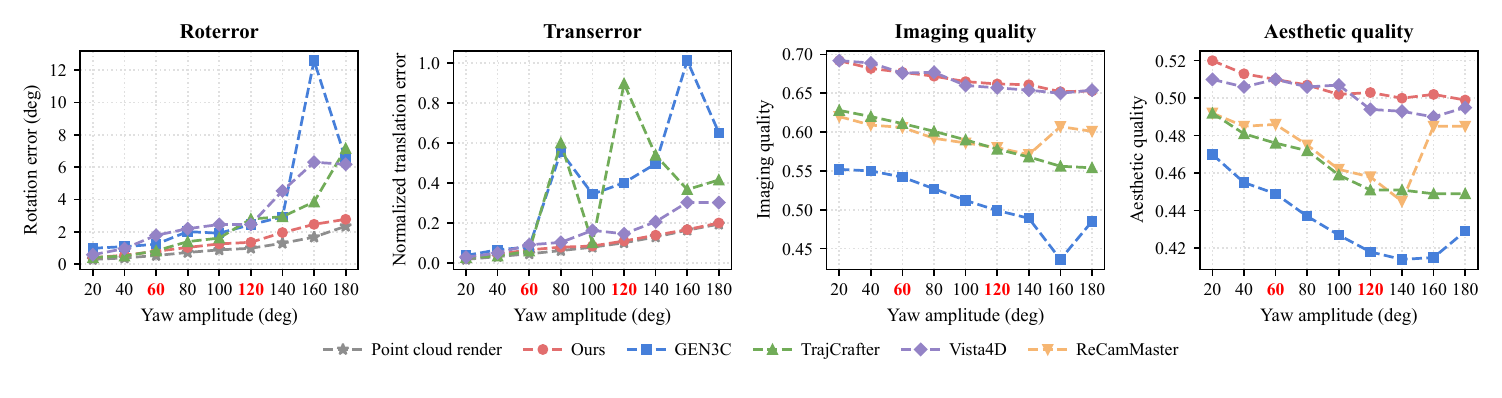}
  \caption{\textbf{Control error and visual quality under growing motion amplitude.} \method{}
maintains the lowest pose error at almost every amplitude, staying close to the
point cloud render, while other methods degrade markedly. Visual quality
(Aesthetic and Imaging) likewise remains the highest, especially at large amplitudes.
Red tick labels mark the amplitudes of \bench{}.}
  \label{fig:yaw_sweep}
\end{figure}

\begin{table}[t]
\centering
\caption{\textbf{Novel view photometric quality on iPhone Dataset.}
\method{} achieves state-of-the-art performance across key photometric and optical-flow metrics.}
\label{tab:iphone_ref}
\footnotesize
\setlength{\tabcolsep}{4pt}
\renewcommand{\arraystretch}{0.9}
\begin{tabular}{lccccccc}
\toprule
Method & PSNR\,$\uparrow$ & SSIM\,$\uparrow$ & LPIPS\,$\downarrow$
       & mPSNR\,$\uparrow$ & mSSIM\,$\uparrow$ & mLPIPS\,$\downarrow$ & EPE\,$\downarrow$ \\
\midrule
ReCamMaster       &           10.89 &           0.3363 &           0.7161 &           11.00 &           0.3326 &           0.5376 &            6.66 \\
TrajectoryCrafter &           15.52 &  \best{0.4338} &           0.5162 &  \second{16.41} &  \best{0.4476} &           0.3494 &            1.58 \\
GEN3C             &           14.75 &  \second{0.3937} &           0.4609 &           15.87 &  \second{0.4119} &           0.3103 &            1.47 \\
Vista4D           &  \second{15.78} &           0.3913 &  \second{0.3884} &           16.30 &           0.3997 &  \second{0.2600} &  \second{1.28} \\
\midrule
\textbf{\ours{}}           &   \best{16.21} &           0.3847 &   \best{0.3582} &   \best{16.65} &           0.3887 &   \best{0.2393} &    \best{1.25} \\
\bottomrule
\end{tabular}
\end{table}

\begin{table}[t]
\centering
\caption{\textbf{Visual quality.} FID/FVD are computed against the corresponding source
clips; Aesthetic and Imaging are VBench scores ($\times100$).}
\label{tab:main_quality}
\footnotesize
\setlength{\tabcolsep}{3pt}
\renewcommand{\arraystretch}{0.9}
\resizebox{\linewidth}{!}{%
\begin{tabular}{lcccccccc}
\toprule
\multirow{2}{*}{Method} & \multicolumn{4}{c}{\bench{}} & \multicolumn{4}{c}{Vista4D-Eval} \\
\cmidrule(lr){2-5} \cmidrule(lr){6-9}
 & FID\,$\downarrow$ & FVD\,$\downarrow$ & Aesthetic\,$\uparrow$ & Imaging\,$\uparrow$ & FID\,$\downarrow$ & FVD\,$\downarrow$ & Aesthetic\,$\uparrow$ & Imaging\,$\uparrow$ \\
\midrule
ReCamMaster       &            65.55 &    \best{1122.9} &           48.099 &           57.204 &     \best{92.79} &    \best{1181.1} &           52.338 &           58.694 \\
TrajectoryCrafter &            85.27 &           1571.6 &           46.899 &           57.038 &           116.31 &           1581.4 &           47.579 &           63.300 \\
GEN3C             &            89.76 &           1463.9 &           47.245 &           53.218 &           113.43 &           1534.3 &           48.852 &           62.001 \\
Vista4D           &     \best{55.64} &  \second{1208.2} &  \second{51.096} &  \second{64.914} &            98.12 &           1404.6 &    \best{55.012} &    \best{70.933} \\
\midrule
\ours{}           &   \second{57.55} &           1222.7 &    \best{51.138} &    \best{65.384} &   \second{97.37} &  \second{1360.2} &  \second{53.757} &  \second{70.509} \\
\bottomrule
\end{tabular}}
\end{table}

\begin{figure}[t]
  \centering
  \includegraphics[width=0.9\linewidth]{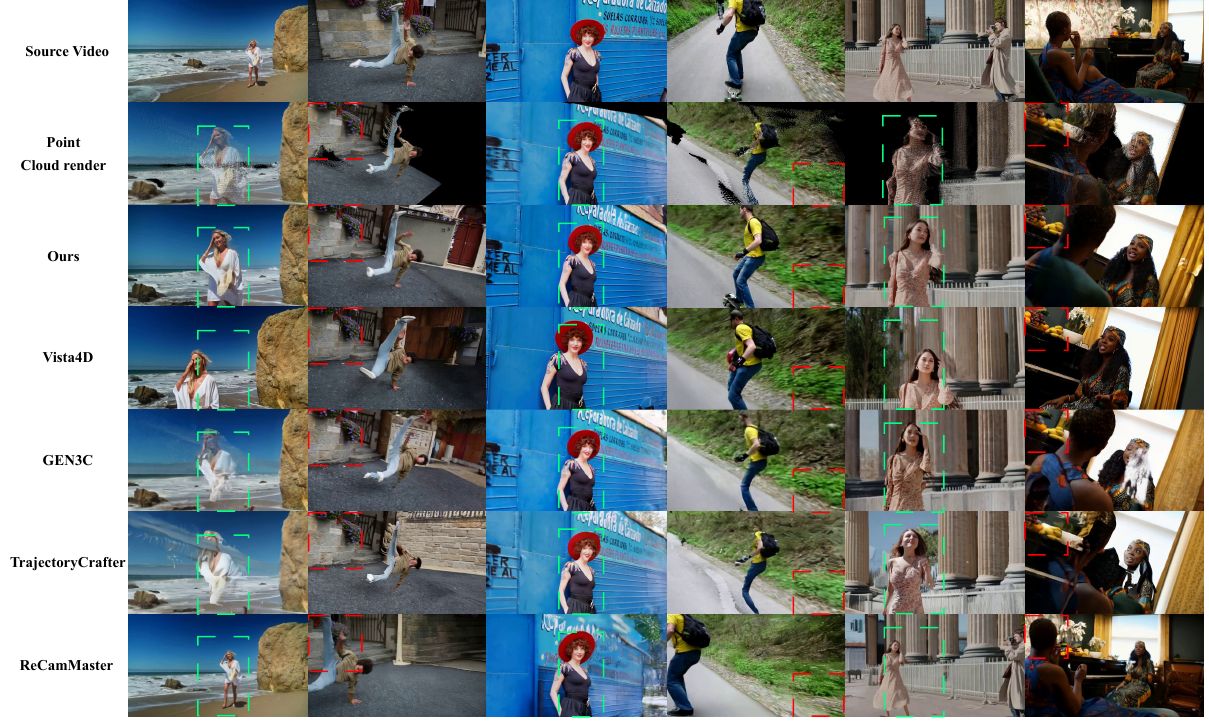}
  \caption{\textbf{Qualitative results on monocular input.}
  Dashed guides are placed on the point cloud render and copied to every row: green
  marks the expected position of the dynamic subject, red a static background structure.
  Our method shows stable control and pleasing visual quality while Vista4D's dynamic subject drifts away from the green guide.
  GEN3C and TrajectoryCrafter inherit the sparsity of the render, leaving
  subjects fragmented; ReCamMaster stays clean but ignores the requested view.
  }
  \label{fig:qual}
\end{figure}

\begin{table}[t]
\centering
\caption{\textbf{User study.} Selection rate in \%. \method{} is preferred by large
margins on trajectory following and 3D consistency of the dynamic subject, and remains
competitive on overall quality.}
\label{tab:user_study}
\footnotesize
\setlength{\tabcolsep}{4pt}
\renewcommand{\arraystretch}{0.9}
\begin{tabular}{lccccc}
\toprule
Criterion & ReCamMaster & TrajectoryCrafter & GEN3C & Vista4D & \ours{} \\
\midrule
Trajectory following & 1.4\% & 17.0\% & 20.5\% & \second{28.3\%} & \best{63.6\%} \\
3D consistency  & 1.8\% & \second{15.6\%} & 9.5\% & 8.7\% & \best{62.4\%} \\
Overall quality      & 12.9\% & 3.2\%  & 5.8\%  & \best{54.7\%} & \second{49.6\%} \\
\bottomrule
\end{tabular}
\end{table}
\begin{table}[t]
\centering
\caption{\textbf{Inference ablation} on Vista4D-Eval.
Pose errors on the 103-scene filtered subset; quality averaged over all 110 scenes.
Left: conditioning ablation. Middle: residual noise strength $\sigma$ sweep.
Right: architecture ablation; all variants share the same base model, seed, and early-stage training budget.}
\label{tab:ablation_infer}
\footnotesize
\setlength{\tabcolsep}{4pt}
\renewcommand{\arraystretch}{0.9}
\resizebox{\textwidth}{!}{%
\begin{minipage}[t]{0.48\textwidth}
\centering
\textbf{(a) Conditioning ablation}\\[2pt]
\begin{tabular}{@{}lcccc@{}}
\toprule
\multirow{2}{*}{Setting} & \multicolumn{2}{c}{Camera} & \multicolumn{2}{c}{Quality} \\
\cmidrule(lr){2-3} \cmidrule(lr){4-5}
 & Rot\,$\downarrow$ & Trans\,$\downarrow$ & AQ\,$\uparrow$ & IQ\,$\uparrow$\\
\midrule
\ours{} (full)          & \second{2.79} & \second{0.374} & 54.0 & \second{70.5} \\
\midrule
w/o camera emb.         & \best{2.62} & \best{0.370} & \second{54.1} & 70.5 \\
w/o text prompt         & 3.86 & 0.458 & 50.0 & 70.0 \\
w/o source video        & 13.07 & 0.636 & 50.4 & 67.3 \\
w/o injected render     & 34.11 & 0.618 & \best{56.5} & \best{71.9} \\
\bottomrule
\end{tabular}
\end{minipage}\hfill
\begin{minipage}[t]{0.32\textwidth}
\centering
\textbf{(b) Residual noise strength $\sigma$}\\[2pt]
\begin{tabular}{@{}lcccc@{}}
\toprule
\multirow{2}{*}{$\sigma$} & \multicolumn{2}{c}{Camera} & \multicolumn{2}{c}{Quality} \\
\cmidrule(lr){2-3} \cmidrule(lr){4-5}
 & Rot\,$\downarrow$ & Trans\,$\downarrow$ & AQ\,$\uparrow$ & IQ\,$\uparrow$\\
\midrule
0.0 & \best{2.71} & \second{0.397} & \second{53.8} & 69.5 \\
0.3 & \second{2.79} & \best{0.374} & \best{54.0} & 70.5 \\
0.6 & 2.90 & 0.407 & 53.5 & \second{71.0} \\
0.9 & 3.89 & 0.461 & 52.3 & \best{71.4} \\
1.2 & 5.73 & 0.545 & 47.3 & 69.4 \\
1.5 & 8.05 & 0.645 & 40.0 & 64.9 \\
\bottomrule
\end{tabular}
\end{minipage}\hfill
\begin{minipage}[t]{0.42\textwidth}
\centering
\textbf{(c) Architecture ablation}\\[2pt]
\begin{tabular}{@{}lcccc@{}}
\toprule
\multirow{2}{*}{Architecture} & \multicolumn{2}{c}{Camera} & \multicolumn{2}{c}{Quality} \\
\cmidrule(lr){2-3} \cmidrule(lr){4-5}
 & Rot\,$\downarrow$ & Trans\,$\downarrow$ & AQ\,$\uparrow$ & IQ\,$\uparrow$\\
\midrule
\ours{} & \best{3.10} & \best{0.391} & 53.7 & \second{69.6} \\
\midrule
Vista4D arch   & \second{3.33} & \second{0.438} & \second{54.2} & \best{70.8} \\
GEN3C arch     & 4.96 & 0.448 & \best{54.7} & 69.0 \\
\bottomrule
\end{tabular}
\end{minipage}
}
\end{table}

\subsection{Experimental setup}
\label{sec:exp:setup}

\paragraph{Benchmarks.}
We select 24 dynamic scenes from DAVIS~\citep{pont2017davis} to form our
benchmark, DAVIS-Traj. Each scene is re-shot under three trajectory patterns
that orbit the dynamic subject while keeping it on the optical axis, sweeping up
to $\pm60$\dg\ of yaw with additional pitch and distance variations, giving 72
evaluation clips; each clip is 49 frames long, and
Fig.~\ref{fig:traj_vis} in Appendix~\ref{app:bench} visualizes the three
patterns and their exact parameterization.
We also select ten of these scenes to construct a dataset of growing
motion amplitudes, probing the limits of model capability.
We additionally report on the 110 evaluation clips released with
Vista4D~\citep{vista4d}, and on the real iPhone multi-view
dataset~\citep{iphone}.
\paragraph{Baselines.}
We compare against four published methods spanning both ways of supplying camera
control: ReCamMaster~\citep{recammaster}, which conditions on an implicit camera
embedding without explicit geometry; TrajectoryCrafter~\citep{yu2025trajectorycrafter}
and GEN3C~\citep{ren2025gen3c}, which concatenate a point-cloud render along the
channel axis; and Vista4D~\citep{vista4d}, which concatenates it at the
token level.
For fairness, every method with explicit geometric control receives the same 4D
point-cloud projection at inference time.
Each method otherwise runs under its own best default settings, and all outputs
are resampled to $384\!\times\!672$ and 49 frames for evaluation.
ReCamMaster and GEN3C output more than 49 frames and require coincident
first-frame poses of the source and target trajectories, so we prepend a
transition segment from the source start to the target start and evaluate the
subsequent 49 frames, which follow the target trajectory.
The base model of every method and its per-clip inference time are reported in
Appendix~\ref{app:inference}.
\paragraph{Point cloud render.}
\label{sec:exp:render}
The point cloud render rasterizes the source point cloud along the target
trajectory, with no generation involved.
We use VGGT-Omega both to reconstruct the point cloud from the
source video and, for evaluation, to estimate poses on the source and each
generated video jointly.
We also evaluate the trajectory of the point cloud render itself 
to quantify the systematic error of the estimator and our alignment protocol; 
the render is excluded from ranking and, since its holes introduce some error, 
treated as a reference rather than strict ground truth.
\subsection{Quantitative comparisons}
\label{sec:exp:main}
\paragraph{Camera control accuracy.}
\label{sec:exp:pose accuracy}
We measure rotation, translation, and intrinsic errors on both benchmarks,
DAVIS-Traj and Vista4D-Eval; the alignment protocol and metric definitions are
given in Appendix~\ref{app:pose-protocol}.
Because some Vista4D-Eval clips have only a small overlap between the source
and target trajectories, coordinate alignment goes conspicuously wrong on them,
so for fairness we
additionally report results after removing clips with a rotation error above
$60$\dg\ in any method or seed.
In Table~\ref{tab:main_pose}, \method{} keeps the lowest rotation and
translation errors in both the filtered and unfiltered settings, and on
rotation even approaches or surpasses the render itself.
The render is weaker on Vista4D-Eval because some scenes contain large empty regions that degrade the pose estimator.
To check that the conclusions do not depend on the pose estimator, we
repeat the pose estimation and alignment with Pi3~\citep{pi3}, and
\method{} again performs best overall; see Appendix~\ref{app:pi3-pose}.
All metrics are averaged over three seeds to reduce random estimation error.

\paragraph{Generalization to larger camera motions.}
\label{sec:exp:magnitude}
We select ten of the benchmark scenes and, for each, construct a family of
trajectories orbiting the dynamic subject, with a per-side motion amplitude
of $10$\dg\ to $90$\dg\ in yaw (total sweep $20$\dg\ to $180$\dg).
We evaluate all methods on this family, together with the
point cloud render as a reference. As Fig.~\ref{fig:yaw_sweep} shows, 
\method{} keeps the lowest rotation error at almost every amplitude, 
and the lowest translation error from moderate amplitudes onward, 
whereas the error of every other method rises markedly.
\method{} also maintains visual quality at large amplitudes,
showing that our explicit design stays stable at amplitudes unseen
during training, while other methods begin to degrade.
ReCamMaster's pose error is too large to plot alongside the other methods.
The complete numerical results are reported in
Tables~\ref{tab:yaw_sweep_pose} and~\ref{tab:yaw_sweep_quality}
of Appendix~\ref{app:yaw-sweep-detail}.
We recommend watching the supplementary demo video for visual results at the largest amplitudes.

\paragraph{Novel-view photometric quality.}
We evaluate photometric quality and 3D consistency on 
the real-world time-synchronized multiview dataset, iPhone~\citep{iphone}.
Table~\ref{tab:iphone_ref} reports PSNR, SSIM~\citep{ssim},
LPIPS~\citep{lpips} and their masked variants
(mPSNR / mSSIM / mLPIPS) over the coverage mask of the point cloud render
(Sec.~\ref{sec:exp:render}),
together with end-point error (EPE) from RAFT~\citep{raft} optical flow, which
measures how closely the generated video follows the ground-truth motion.
\method{} performs best on all metrics except SSIM and mSSIM, where
TrajectoryCrafter leads.
However, viewing the synthesized videos quickly reveals significant
artifacts in TrajectoryCrafter's outputs that SSIM fails to capture.
We show examples in Fig.~\ref{fig:iphone_qual} of Appendix~\ref{app:results}.

\paragraph{Video fidelity.}
\label{sec:exp:video fidelity}
We evaluate video fidelity and quality on both benchmarks, DAVIS-Traj and
Vista4D-Eval, reporting FID~\citep{fid}, FVD~\citep{fvd} and
VBench~\citep{vbench} scores (aesthetic quality and
imaging quality) in Table~\ref{tab:main_quality}; the remaining metrics are reported in Appendix~\ref{app:full-quality}. 
\method{} is on par with Vista4D across the board,
which means the tight trajectory control does not come at the cost of visual quality through copying the render.
ReCamMaster scores best on FVD on both benchmarks (and on FID on Vista4D-Eval), which we attribute to
its tendency to ignore the target trajectory and stay close to the source
content.
The low quality of GEN3C and TrajectoryCrafter stems from over-reliance on the render,
which fragments and blurs the dynamic content.
We show examples in Fig.~\ref{fig:qual}.
\subsection{Qualitative comparisons}
\label{sec:exp:qual}
\paragraph{Geometric alignment of dynamic content.}
\label{sec:exp:3d consistency}
In Fig.~\ref{fig:qual}, the red and green dashed guides track the static
background and the expected position of the dynamic subject, respectively,
probing 3D consistency on both.
Vista4D keeps the static background aligned with the render, but its dynamic subject drifts away from the green guide.
This behavior also explains its low trajectory error: for pose estimation,
only the static background provides reliable geometric cues.
GEN3C and TrajectoryCrafter are similarly consistent, but when sparsity
lets the background leak through the point cloud, they fail to complete the
missing content and leave the subject fragmented.
ReCamMaster, in contrast, loses geometric control altogether.
Taken together, \method{} simultaneously delivers precise camera control, dynamic-object consistency, 
and visual quality, and exhibits the strongest stability at the boundary of its capability.
We also provide a supplementary demo video with additional examples.

\paragraph{User study.}
We present the results of our user study in Table~\ref{tab:user_study}, where we
ask users to evaluate all methods in three aspects: trajectory following, 3D consistency of the dynamic subject, and visual quality (multiple selections are permitted).
Twenty clips, ten sampled from each of the two benchmarks, are rated by over 30
participants, who do not know the generating model.
The results show \method{} achieves a clear advantage in trajectory
following and 3D consistency of the dynamic subject, while ranking second on visual quality
with a narrow margin to Vista4D.
To help participants judge objectively, we provide an optional overlay of
the point cloud render, making geometric misalignment directly visible; the
evaluation interface is shown in Appendix~\ref{app:user-study-interface}.

\paragraph{Robustness to imperfect geometry.}
\label{sec:exp:robustness}
To test how \method{} copes with a degraded point cloud, we remove the
motion mask at render time so that all dynamic points are stacked together
across frames, producing a render with incorrect dynamic geometry.
Despite the corrupted render, the model still recovers the correct motion
from the source video.
This highlights the benefit of injecting the render only once: errors in the
render can be corrected from the source video during denoising.
Visualizations are shown in Fig.~\ref{fig:rb_test},
Appendix~\ref{app:robustness}.

\subsection{Ablation study}
\label{sec:exp:ablation}
\paragraph{Inference-time ablation.}
\label{sec:exp:abl-infer}
Table~\ref{tab:ablation_infer}(a) removes each conditioning stream at inference time.
Removing the camera embedding leaves trajectory control essentially unchanged.
The slight pose gain is within the noise of single-seed evaluation rather than
a genuine improvement.
Removing the text prompt causes a mild degradation.
The source video and the point cloud render are both critical:
suppressing the source video degrades rotation by $4.7\times$, and suppressing
the render collapses it by $12\times$.
Note that \emph{w/o injected render} achieves the highest visual quality,
as it degenerates into replicating the source video.
Table~\ref{tab:ablation_infer}(b) sweeps the residual noise strength~$\sigma$
in Eq.~\ref{eq:init}, the extra noise overlaid on the render at the starting
point, from the noise-free render up to $5\times$ the default.
Moderate residual noise improves imaging quality, while excessive noise
degrades both control and quality.
The noise-free variant ($\sigma{=}0$) retains the strongest camera control
but lower visual quality.
Starting directly from the raw render leaves the model no room to correct
it, while too much noise drowns out the geometric prior.
Moderate noise instead provides this freedom, so the model can refine the
render and recover the details it misses.
We adopt $\sigma{=}0.3$ as the default,
which attains the best translation accuracy and aesthetic quality.

\paragraph{Architecture variants.}
To separate the contributions of our architecture and our training recipe,
we train the conditioning layouts of the two strongest baselines with the
same base model, seed and early-stage training budget.
In Table~\ref{tab:ablation_infer}(c), both variants fall behind \method{} on
camera control, with visual quality comparable across the three.
The Vista4D-style variant trades pose accuracy for a small imaging-quality gain,
and also shows lingering geometric inconsistency on dynamic objects (milder than
the original Vista4D) and about $1.5\times$ the inference time.
The GEN3C-style variant shows visible quality degradation in some scenes despite
a slightly higher aesthetic score, and its rotation error is $1.6\times$ ours.
The advantage of \method{} therefore comes from architecture and training
recipe together.
In this controlled comparison, the simplest layout wins.
Appendix~\ref{app:abl} visualizes the three layouts side by side, making the
artifact patterns directly visible.

\section{Conclusion}
\label{sec:conclusion}

We have presented \method{}, which re-shoots a monocular video along a
user-specified camera trajectory.
By injecting explicit geometry into the noise once and keeping the
source video as the only visual condition, \method{} resolves the competition
between conditions and uses the geometric prior directly.
Extensive experiments show that this simple design suffices: \method{} attains
the best trajectory control on every benchmark with competitive visual quality,
stays geometrically consistent on dynamic objects and reliable under large
camera motions.

\paragraph{Limitations and future work.}
We believe the generative prior of the pretrained diffusion model still holds
potential that more sophisticated noise--geometry interactions could unlock.
Designing such interactions is a promising direction for future work.

\bibliographystyle{iclr2027_conference}
\bibliography{refs}

\appendix
\section{Teaser figure details}
\label{app:details}
In the teaser (Fig.~\ref{fig:teaser}), each method is plotted on both
benchmarks from its metrics averaged over three seeds.
The horizontal axis gives the camera-control error relative to the best
method: RotErr and TransErr, computed on the
common scene set, are each divided by the lowest value among the compared
methods on that benchmark, and the two ratios are averaged, so the best
method sits at $1$; the axis uses a base-2 logarithmic scale.
The vertical axis averages the two VBench scores, aesthetic and imaging
quality, from Table~\ref{tab:main_quality} ($\times100$).

\section{Training details}
\label{app:train}

\paragraph{Base model and trainable subset.}
We build on Wan2.1-T2V-14B together with its
native 16-channel VAE, which downsamples by $8\times$ spatially and $4\times$
temporally.
Clips are 49 frames at $384\!\times\!672$, mapped by the VAE to 13 latent frames.
Only the self-attention projections (Q/K/V/O with their RMSNorm layers), the
four patchify embeddings, the per-block projectors applied to the
self-attention output, and the per-block Pl\"ucker camera encoder are trainable;
feed-forward blocks,
cross-attention, normalization layers, time and text embeddings, the output head
and all modulation parameters remain frozen.
The camera encoder is
zero-initialized, so at step $0$ the network is bit-identical to the
pretrained backbone and the geometric pathway grows in from zero.

\paragraph{Training data.}
We train in a single mixed stage over five sources: DL3DV~\citep{ling2024dl3dv}
(8.5k static scenes), DynPose~\citep{dynpose100k} (8.0k) and
OpenVid-HD~\citep{openvid} (14.9k) monocular
dynamic clips, MultiCamVideo~\citep{recammaster} (3.2k synthetic scenes with ten
cameras each, captured at three focal lengths), and HuMMan~\citep{cai2022humman}
(0.9k real ten-camera human captures).
In our experience DL3DV and HuMMan matter comparatively little for final quality.
Sampling weights give MultiCamVideo $56\%$, the monocular dynamic clips $22\%$, DL3DV
$11\%$ and HuMMan $11\%$ of each epoch.
Metric depth and camera poses come from VGGT-Omega~\citep{vggtomega} for every source except
HuMMan, which ships with calibrated multi-camera captures.
The monocular clips carry no second view, so they are used through a
double-reprojection cycle~\citep{yu2025trajectorycrafter}: a synthetic
target camera is sampled (up to $15$\dg\ yaw and $0.30$ of the scene
extent in translation, rejected unless the warp retains at least $45\%$
coverage), the source is reprojected into it and back, and the round
trip supplies a supervised pair.

\paragraph{Optimization.}
We train for 30k steps with AdamW at a constant learning rate of $10^{-5}$, batch
size $1$ per GPU on eight A100s.
Each conditioning stream (source, point cloud, camera, prompt) is dropped
independently with probability $10\%$; dropping the point cloud sets the coverage
$\alpha$ to zero, so the starting state falls back to pure Gaussian noise and the
sample reverts to standard flow matching, and a further $5\%$
of samples are trained fully unconditionally to preserve the text-to-video prior.
To prevent the model from copying the warp where it is unreliable, we augment the
point cloud at render time: a slab of the dynamic subject is carved out with
probability $20\%$, and static points are removed in spatial blocks or along
temporal spans, each with probability $20\%$.
Inspired by Vista4D~\citep{vista4d}, MultiCamVideo and HuMMan additionally use
$50\%$ temporal reversal.

\section{Benchmark construction details}
\label{app:bench}
\begin{figure}[t]
  \centering
  \includegraphics[width=\linewidth]{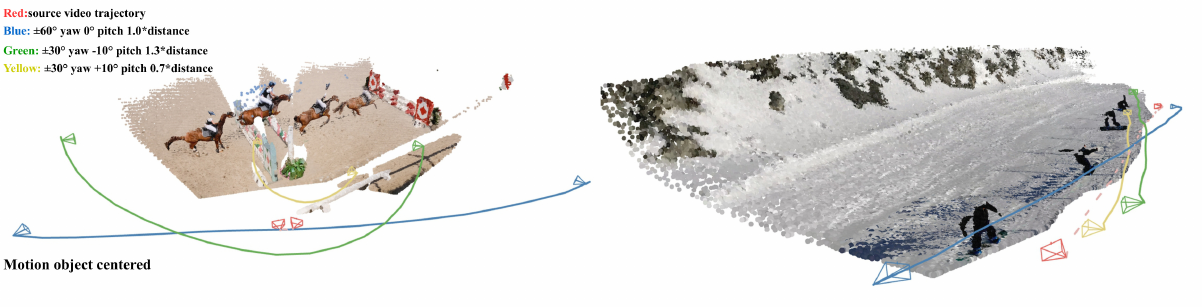}
  \caption{\textbf{Trajectory patterns of \bench{}.} Each pattern orbits the
  dynamic subject while keeping it on the optical axis. f1 (blue) sweeps
  $\pm60$\dg\ of yaw; f2 (green) sweeps $\pm30$\dg\ of yaw with $-10$\dg\ of
  pitch at $1.3\times$ distance; f3 (yellow) sweeps $\pm30$\dg\ of yaw with
  $+10$\dg\ of pitch while dollying in from $1.0$ to $0.7$. The red curve
  shows the source camera trajectory.}
  \label{fig:traj_vis}
\end{figure}

\paragraph{Scene selection and trajectory patterns.}
We select 24 dynamic scenes from DAVIS~\citep{pont2017davis} and re-shoot each
of them under three trajectory patterns (f1--f3), giving 72 evaluation clips; a
representative sample is visualized in Fig.~\ref{fig:traj_vis}.
We additionally report on the 110 evaluation clips released with
Vista4D~\citep{vista4d}, so that every method is scored on both a purpose-built
and a third-party benchmark.
All three patterns orbit the dynamic subject while keeping it on the optical
axis.
Each trajectory is parameterized by a per-frame (yaw, pitch, dolly) triple,
interpolated linearly over the 49 frames of the clip:
\begin{itemize}
  \item \textbf{f1} sweeps yaw from $-60^{\circ}$ to $+60^{\circ}$ at the source camera
  distance (pitch $0^{\circ}$, dolly $1.0$);
  \item \textbf{f2} sweeps yaw from $+30^{\circ}$ to $-30^{\circ}$ at a fixed pitch of
  $-10^{\circ}$ and $1.3\times$ the source camera distance;
  \item \textbf{f3} sweeps yaw from $+30^{\circ}$ to $-30^{\circ}$ at a fixed pitch of
  $+10^{\circ}$ while dollying in from $1.0\times$ to $0.7\times$ the source camera
  distance.
\end{itemize}

\paragraph{Alignment conditioning.}
Each pattern sweeps the camera through the source viewpoint.
The mid-clip frames pass close to the source camera, and the generated video
retains a large overlap with the source video.
This overlap anchors the estimated trajectory to the source content and keeps
the $\mathrm{Sim}(3)$ alignment of Appendix~\ref{app:pose-protocol} well
conditioned.
\bench{} therefore does not exhibit the degenerate alignments removed by the
filtering described there.

\paragraph{Growing-magnitude trajectories.}
For the generalization study of Sec.~\ref{sec:exp:magnitude}, we construct
symmetric yaw sweeps on ten of the benchmark scenes, with per-side amplitudes
from $10^{\circ}$ to $90^{\circ}$ (total sweep $20^{\circ}$--$180^{\circ}$) at zero pitch and unit
dolly.

\section{Evaluation protocol}
\label{app:pose-protocol}

\paragraph{Trajectory alignment.}
For each generated clip, we run VGGT-Omega~\citep{vggtomega} on the source and
generated videos jointly, so their estimated cameras live in one coordinate
frame. Given the estimated source cameras $\{(R^e_i, C^e_i)\}$ and their
ground-truth counterparts $\{(R^g_i, C^g_i)\}$, we recover a global similarity
transform $(s, R, \mathbf{t})$ that maps the estimated source trajectory to
ground truth.
The rotation $R$ is obtained by solving the orthogonal Procrustes problem on
the camera orientations: we form the correlation matrix
$M = \sum_i R^g_i (R^e_i)^\top$ and take its SVD $M = U \Sigma V^\top$; the
optimal rotation is $R = U D V^\top$ where $D = \mathrm{diag}(1, 1, \det(UV^\top))$
guards against reflections.
This orientation-based recovery is more robust than center-only SVD when the
camera path is near-planar or short.
The scale $s$ and translation $\mathbf{t}$ are then the least-squares fit of the
camera centers given that rotation:
$C^g \approx s\, R\, C^e + \mathbf{t}$.
Since the generated cameras share the coordinate frame of the estimated
source cameras, the same transform brings them into the ground-truth world,
and we can compare them directly with the requested target trajectory,
following the protocol described by \citet{vista4d}.

\paragraph{Metrics.}
We report three pose metrics, all computed per-frame and averaged over $T$
generated frames:
\begin{gather}
\text{RotErr} = \frac{1}{T}\sum_{i=1}^{T}
\arccos\!\left(\frac{\mathrm{tr}(R^{\text{tgt}\top}_i R^{\text{gen}}_i) - 1}{2}\right), \\
\text{TransErr} = \frac{1}{\bar{d}} \sqrt{\frac{1}{T}\sum_{i=1}^{T}
\|C^{\text{tgt}}_i - C^{\text{gen}}_i\|^2}, \\
\text{IntrErr} = \frac{1}{T}\sum_{i=1}^{T}
\bigl|\text{FOV}_v(f^{\text{tgt}}_i) - \text{FOV}_v(f^{\text{gen}}_i)\bigr|,
\end{gather}
where $\bar{d}$ is the median source scene depth (normalizing by $\bar{d}$
removes the bias from each dataset's arbitrary world scale, enabling fair
cross-dataset comparison), $\text{FOV}_v(f) = 2\arctan\!\bigl(\tfrac{1}{2f}\bigr)$
in degrees, and $f$ is the height-normalized focal length ($f_y / H$).

\paragraph{Common clip set and degenerate alignment filtering.}
All methods are evaluated on the maximum common clip set.
We remove clips where RotErr $> 60^\circ$ in any method or seed,
indicating a catastrophically wrong alignment rather than a failure of the
generated video itself (we verified this through manual inspection).
These failures occur exclusively on the Vista4D-Eval benchmark, where
across all methods and seeds, 7/110 clips trigger this filter, typically when the
source camera is nearly static and the overlap between source and target views
is so small that, after source alignment, the target trajectory becomes
mirrored relative to the ground truth.
Some clips that escape filtering still suffer from elevated errors due to the
same low-overlap pathology, though to a lesser degree.
Our proposed benchmark (\bench{}) does not exhibit this issue
(Appendix~\ref{app:bench}).
This filtering ensures that the reported numbers reflect genuine camera control
quality rather than reconstruction failures of the pose estimator
on pathological inputs, and the maximum-common-set constraint guarantees that
all methods are scored on identical clips so no method benefits from a
selective evaluation subset.

\section{Additional quantitative results}
\label{app:quant-results}

\subsection{Inference time}
\label{app:inference}
\begin{table}[t]
\centering
\caption{\textbf{Base models and inference time.} All methods run with 50 diffusion
steps on a single NVIDIA A100 80\,GB GPU and generate at their native resolution.}
\label{tab:inference_time}
\footnotesize
\setlength{\tabcolsep}{4pt}
\begin{tabular}{llcl}
\toprule
Method & Base model & Output & Time (min) \\
\midrule
ReCamMaster       & Wan2.1-T2V-1.3B   & 81$\times$480$\times$832   &   9.0 \\
TrajectoryCrafter & CogVideoX-Fun 5B  & 49$\times$384$\times$672   &   2.7 \\
GEN3C             & Cosmos-Predict1 7B & 121$\times$704$\times$1280 &  18.5 \\
Vista4D           & Wan2.1-T2V-14B    & 49$\times$384$\times$672   &  19.3 \\
\cmidrule(lr){1-4}
\ours{}           & Wan2.1-T2V-14B    & 49$\times$384$\times$672   &  13.0 \\
\bottomrule
\end{tabular}
\end{table}

Table~\ref{tab:inference_time} compares the base models and per-clip inference
time of all methods, measured on the same \bench{} clips with 50 diffusion
steps on a single NVIDIA A100 80\,GB GPU.
The baselines differ in base model and output format, so absolute times are
not directly comparable.
Vista4D is the only baseline that uses the same base model as \method{}.
Compared with Vista4D, \method{} is about $1.5\times$ faster and achieves
better camera control.

\subsection{Pi3 pose estimation results}
\label{app:pi3-pose}
\begin{table}[t]
\centering
\caption{\textbf{Camera control accuracy with Pi3 reconstruction.}
\emph{Left}: filtered (RotErr~$\leq60^\circ$ in any method, maximum common
clip set). \emph{Right}: raw Vista4D-Eval (all 110 clips, no filtering).
\bench{} has no filtered clips so only one column group is shown.
3-seed mean. \best{Bold} denotes the best and \second{underline} denotes
the second best among generated methods (Point cloud render excluded from
ranking and typeset in italics).}
\label{tab:pi3_pose}
\small
\setlength{\tabcolsep}{4pt}
\resizebox{\textwidth}{!}{
\begin{tabular}{lccccccccc}
\toprule
\multirow{2}{*}{Method} & \multicolumn{3}{c}{\bench{} (72 clips)}
 & \multicolumn{3}{c}{Vista4D-Eval (106 clips)}
 & \multicolumn{3}{c}{Vista4D-Eval raw (110 clips)} \\
\cmidrule(lr){2-4}\cmidrule(lr){5-7}\cmidrule(lr){8-10}
 & RotErr\,$\downarrow$ & TransErr\,$\downarrow$ & IntrErr\,$\downarrow$
 & RotErr\,$\downarrow$ & TransErr\,$\downarrow$ & IntrErr\,$\downarrow$
 & RotErr\,$\downarrow$ & TransErr\,$\downarrow$ & IntrErr\,$\downarrow$ \\
\midrule
ReCamMaster & 14.00 & 0.586 & 3.19 & 13.23 & 0.549 & 10.67
 & 13.56 & 0.563 & 10.67 \\
TrajectoryCrafter & \second{1.79} & \second{0.256} & \second{1.52} & 6.62 & 0.500 & 7.55
 & 7.71 & 0.504 & 7.57 \\
GEN3C & 1.88 & 0.267 & \best{1.41} & \second{4.05} & 0.438 & 7.52
 & \best{4.25} & 0.433 & 7.52 \\
Vista4D & 2.90 & 0.304 & 2.06 & 4.07 & \best{0.382} & \second{7.23}
 & 4.67 & \best{0.388} & \second{7.26} \\
\textbf{\ours{}} & \best{1.56} & \best{0.247} & 1.59
 & \best{3.82} & \second{0.419} & \best{6.96}
 & \second{4.39} & \second{0.427} & \best{6.97} \\
\midrule
Point cloud render & \textit{1.37} & \textit{0.226} & \textit{1.44} & \textit{3.62} & \textit{0.438} & \textit{6.74}
 & \textit{6.32} & \textit{0.457} & \textit{6.81} \\
\bottomrule
\end{tabular}}
\end{table}

Table~\ref{tab:pi3_pose} reports camera control accuracy estimated by
Pi3~\citep{pi3} instead of VGGT-Omega.
The left half applies the RotErr~$\leq60^\circ$ filter on the maximum
common clip set; the right half reports raw metrics on all clips with
no filtering.
Since catastrophic alignment failures are estimator-dependent, this
filter is recomputed under the Pi3 estimates and retains 106 of the 110
clips.
The ``Point cloud render'' row is the point-cloud-rendered video (no generation),
included as a geometric reference and excluded from ranking.
Under the Pi3 estimate all methods show inflated errors, but \method{} still
ranks best overall.

\subsection{Detailed yaw-sweep results}
\label{app:yaw-sweep-detail}
\begin{table}[t]
\centering
\caption{\textbf{Camera control accuracy under growing motion amplitude.}
RotErr and TransErr at each amplitude, averaged over ten \bench{} scenes.
The amplitude is the per-side yaw angle, from $10^\circ$ to $90^\circ$
(total sweep $20^\circ$--$180^\circ$).
The point cloud render is included as a geometric reference and excluded from ranking.}
\label{tab:yaw_sweep_pose}
\small
\setlength{\tabcolsep}{3pt}
\renewcommand{\arraystretch}{0.9}
\resizebox{\textwidth}{!}{
\begin{tabular}{lcccccccccccccccccc}
\toprule
& \multicolumn{9}{c}{RotErr\,$\downarrow$ (deg)}
 & \multicolumn{9}{c}{TransErr\,$\downarrow$ (normalized)} \\
\cmidrule(lr){2-10}\cmidrule(lr){11-19}
Yaw & $10$ & $20$ & $30$ & $40$ & $50$ & $60$ & $70$ & $80$ & $90$
 & $10$ & $20$ & $30$ & $40$ & $50$ & $60$ & $70$ & $80$ & $90$ \\
\midrule
ReCamMaster & 7.32 & 9.21 & 11.44 & 14.43 & 17.67 & 19.55 & 22.28 & 38.94 & 43.19
 & 0.290 & 0.348 & 0.510 & 0.622 & 0.560 & 0.689 & 0.721 & 0.918 & 0.853 \\
TrajectoryCrafter & \second{0.41} & \best{0.50} & \second{0.82} & \second{1.39} & \second{1.61} & 2.79 & 2.93 & \second{3.85} & 7.15
 & \best{0.028} & \best{0.035} & \best{0.060} & 0.600 & \second{0.101} & 0.896 & 0.539 & 0.367 & 0.415 \\
GEN3C & 0.98 & 1.08 & 1.23 & 2.01 & 1.91 & \second{2.44} & \second{2.92} & 12.58 & 6.49
 & 0.038 & 0.066 & 0.080 & 0.557 & 0.344 & 0.401 & 0.496 & 1.011 & 0.651 \\
Vista4D & 0.58 & 0.96 & 1.78 & 2.19 & 2.45 & 2.45 & 4.53 & 6.30 & \second{6.17}
 & \second{0.029} & 0.050 & 0.091 & \second{0.103} & 0.163 & \second{0.146} & \second{0.206} & \second{0.303} & \second{0.302} \\
\ours{} & \best{0.40} & \second{0.55} & \best{0.79} & \best{1.03} & \best{1.25} & \best{1.35} & \best{1.95} & \best{2.46} & \best{2.77}
 & 0.034 & \second{0.044} & \second{0.066} & \best{0.079} & \best{0.085} & \best{0.110} & \best{0.139} & \best{0.167} & \best{0.200} \\
\midrule
Point cloud render & \textit{0.29} & \textit{0.38} & \textit{0.53} & \textit{0.72} & \textit{0.88} & \textit{0.98} & \textit{1.29} & \textit{1.67} & \textit{2.34}
 & \textit{0.018} & \textit{0.032} & \textit{0.046} & \textit{0.062} & \textit{0.079} & \textit{0.101} & \textit{0.130} & \textit{0.164} & \textit{0.194} \\
\bottomrule
\end{tabular}}

\vspace{4pt}

\caption{\textbf{Visual quality under growing motion amplitude.}
VBench aesthetic and imaging quality scores at each amplitude,
averaged over ten \bench{} scenes.}
\label{tab:yaw_sweep_quality}
\small
\setlength{\tabcolsep}{3pt}
\renewcommand{\arraystretch}{0.9}
\resizebox{\textwidth}{!}{
\begin{tabular}{lcccccccccccccccccc}
\toprule
& \multicolumn{9}{c}{Aesthetic quality $\uparrow$}
 & \multicolumn{9}{c}{Imaging quality $\uparrow$} \\
\cmidrule(lr){2-10}\cmidrule(lr){11-19}
Yaw & $10$ & $20$ & $30$ & $40$ & $50$ & $60$ & $70$ & $80$ & $90$
 & $10$ & $20$ & $30$ & $40$ & $50$ & $60$ & $70$ & $80$ & $90$ \\
\midrule
ReCamMaster & 0.492 & 0.485 & 0.486 & 0.475 & 0.462 & 0.458 & 0.445 & 0.485 & 0.485
 & 0.620 & 0.609 & 0.606 & 0.592 & 0.586 & 0.580 & 0.571 & 0.607 & 0.601 \\
TrajectoryCrafter & 0.492 & 0.481 & 0.476 & 0.472 & 0.459 & 0.451 & 0.451 & 0.449 & 0.449
 & 0.628 & 0.620 & 0.611 & 0.601 & 0.590 & 0.578 & 0.568 & 0.556 & 0.554 \\
GEN3C & 0.470 & 0.455 & 0.449 & 0.437 & 0.427 & 0.418 & 0.414 & 0.415 & 0.429
 & 0.552 & 0.550 & 0.542 & 0.527 & 0.512 & 0.499 & 0.489 & 0.436 & 0.485 \\
Vista4D & \second{0.510} & \second{0.506} & \best{0.510} & \second{0.506} & \best{0.507} & \second{0.494} & \second{0.493} & \second{0.490} & \second{0.495}
 & \second{0.692} & \best{0.689} & \second{0.676} & \best{0.677} & \second{0.660} & \second{0.657} & \second{0.654} & \second{0.650} & \best{0.654} \\
\ours{} & \best{0.520} & \best{0.513} & \second{0.510} & \best{0.507} & \second{0.502} & \best{0.503} & \best{0.500} & \best{0.502} & \best{0.499}
 & \best{0.692} & \second{0.682} & \best{0.677} & \second{0.672} & \best{0.665} & \best{0.662} & \best{0.661} & \best{0.652} & \second{0.653} \\
\bottomrule
\end{tabular}}
\end{table}

These tables supplement Fig.~\ref{fig:yaw_sweep} of
Sec.~\ref{sec:exp:magnitude} with the full numerical results.
Table~\ref{tab:yaw_sweep_pose} reports the rotation and translation errors
of each method at every amplitude.
Table~\ref{tab:yaw_sweep_quality} reports the corresponding VBench aesthetic
and imaging quality scores.

\subsection{Full quality metrics}
\label{app:full-quality}
\begin{table}[t]
\centering
\caption{\textbf{Full visual quality and temporal stability metrics.} FID/FVD are
computed against the corresponding source clips, CLIP-T is the frame-wise text-image
similarity, and the remaining five columns are VBench scores ($\times100$): subject
consistency (SC), background consistency (BC), motion smoothness (MS), aesthetic quality
(AQ) and imaging quality (IQ). Mean over 3 seeds.}
\label{tab:full_quality}
\small
\setlength{\tabcolsep}{4pt}
\resizebox{\linewidth}{!}{%
\begin{tabular}{lcccccccc}
\toprule
Method & FID\,$\downarrow$ & FVD\,$\downarrow$ & CLIP-T\,$\uparrow$
 & SC\,$\uparrow$ & BC\,$\uparrow$ & MS\,$\uparrow$ & AQ\,$\uparrow$ & IQ\,$\uparrow$ \\
\midrule
\multicolumn{9}{l}{\emph{\bench{}}} \\
ReCamMaster       &            65.55 &    \best{1122.9} &           0.2392 &    \best{88.333} &           89.807 &    \best{97.932} &           48.099 &           57.204 \\
TrajectoryCrafter &            85.27 &           1571.6 &           0.2361 &           84.866 &           88.519 &  \second{95.911} &           46.899 &           57.038 \\
GEN3C             &            89.76 &           1463.9 &           0.2298 &           84.543 &    \best{90.624} &           95.627 &           47.245 &           53.218 \\
Vista4D           &     \best{55.64} &  \second{1208.2} &    \best{0.2508} &  \second{86.980} &  \second{90.420} &           95.677 &  \second{51.096} &  \second{64.914} \\
\cmidrule(lr){1-9}
\ours{}           &   \second{57.55} &           1222.7 &  \second{0.2444} &           86.393 &           89.367 &           95.122 &    \best{51.138} &    \best{65.384} \\
\midrule
\multicolumn{9}{l}{\emph{Vista4D-Eval}} \\
ReCamMaster       &     \best{92.79} &    \best{1181.1} &           0.2478 &    \best{91.295} &    \best{91.656} &    \best{98.886} &           52.338 &           58.694 \\
TrajectoryCrafter &           116.31 &           1581.4 &           0.2317 &           85.335 &           88.032 &           97.672 &           47.579 &           63.300 \\
GEN3C             &           113.43 &           1534.3 &           0.2246 &           85.984 &           90.804 &  \second{97.958} &           48.852 &           62.001 \\
Vista4D           &            98.12 &           1404.6 &    \best{0.2534} &  \second{87.948} &  \second{91.338} &           97.505 &    \best{55.012} &    \best{70.933} \\
\cmidrule(lr){1-9}
\ours{}           &   \second{97.37} &  \second{1360.2} &  \second{0.2498} &           86.948 &           88.649 &           97.511 &  \second{53.757} &  \second{70.509} \\
\bottomrule
\end{tabular}}
\end{table}

Table~\ref{tab:full_quality} shows the full quality metrics on both
benchmarks, where CLIP-T is the frame-wise CLIP~\citep{clip} text-image
similarity and the remaining columns are VBench~\citep{vbench} scores.
ReCamMaster leads on FVD, SC and MS on both benchmarks, and also on FID and
BC on Vista4D-Eval.
We attribute this to its considerably weaker trajectory control.
Our method is on par with Vista4D on these metrics.

\section{Additional qualitative results}
\label{app:results}
\subsection{Visualization quality}
\begin{figure}[t]
  \centering
  \includegraphics[width=\linewidth]{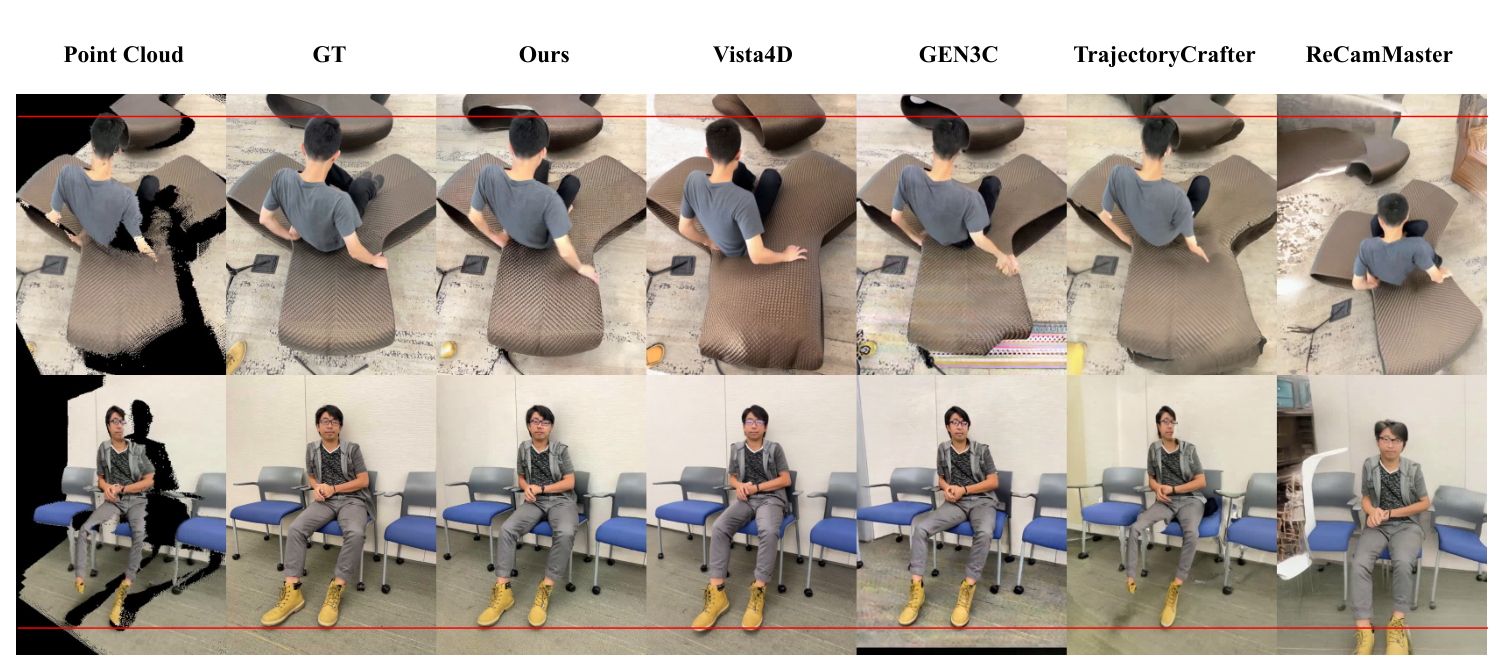}
  \caption{\textbf{Novel-view comparison on real iPhone multi-view captures.}
  Vista4D shows mild 3D inconsistency (red lines) and does not follow the
  point cloud colors; TrajectoryCrafter and GEN3C appear flat and miss fine
  human details; ReCamMaster lacks camera control.}
  \label{fig:iphone_qual}
\end{figure}

Fig.~\ref{fig:iphone_qual} shows qualitative results on the iPhone dataset.
The red guide lines highlight 3D inconsistency: Vista4D exhibits mild
misalignment on the dynamic subject, while GEN3C and TrajectoryCrafter show
visibly worse quality on dynamic objects.
ReCamMaster loses trajectory control entirely and deviates substantially from
the ground truth.

\subsection{Robustness to imperfect geometry}
\label{app:robustness}
\begin{figure}[t]
  \centering
  \includegraphics[width=\linewidth]{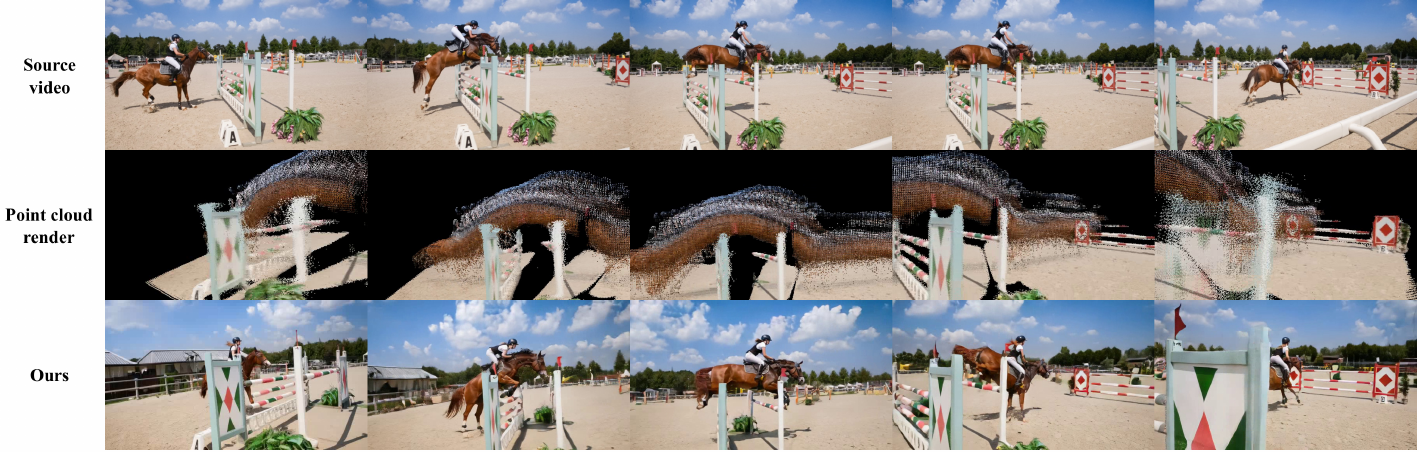}
  \caption{\textbf{Robustness to incorrect dynamic geometry.} We omit the motion mask to simulate an erroneous point cloud: dynamic
  points from all frames are stacked together, smearing the rendered
  subject. \method{} still recovers the correct motion from the source
  video.}
  \label{fig:rb_test}
\end{figure}

Fig.~\ref{fig:rb_test} visualizes the robustness test described in
Sec.~\ref{sec:exp:robustness}.
When the motion mask is removed at render time, all dynamic points are
stacked across frames, producing a render with incorrect dynamic geometry.
Despite this corrupted input, \method{} still recovers the correct motion
from the source video.
We do observe a moderate quality degradation in fine details, particularly
a residual blur where the stacked points overlap, which is expected given
the conflicting geometric signal.

\begin{figure}[t]
  \centering
  \includegraphics[width=\linewidth]{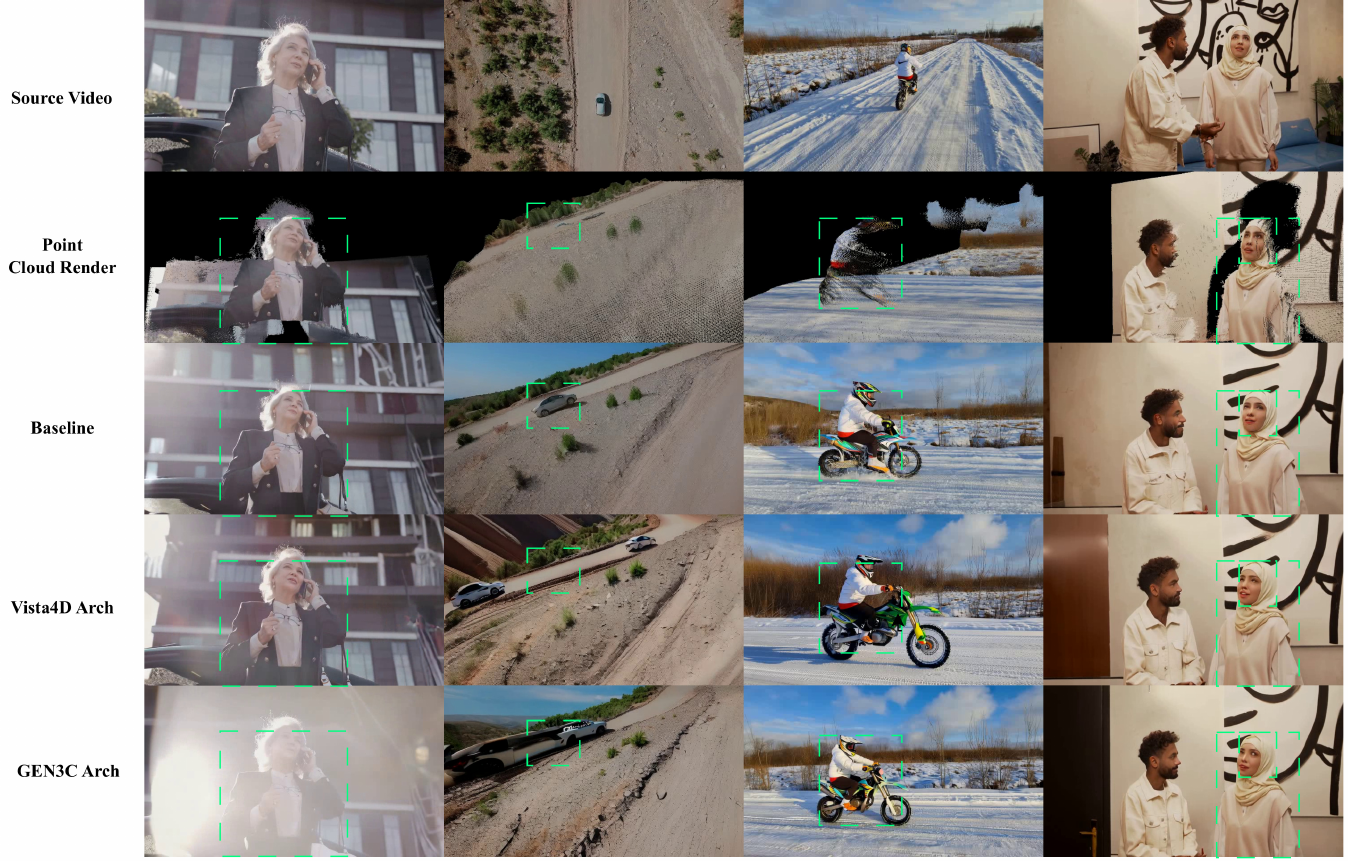}
  \caption{\textbf{Qualitative ablation comparison.} The two architecture
  variants against the full model on Vista4D-Eval clips.
  The Vista4D-style three-stream layout exhibits 3D inconsistency on dynamic
  objects, while the GEN3C-style variant shows noticeable degradation in
  visual quality in some scenes.
  The full model preserves both geometric consistency and visual quality.}
  \label{fig:ablation_qual}
\end{figure}

\subsection{Ablation study on architecture variants}
\label{app:abl}
Fig.~\ref{fig:ablation_qual} provides visualizations for the architecture
ablation of Sec.~\ref{sec:exp:ablation}.
The Vista4D-style three-stream layout exhibits 3D inconsistency on dynamic
objects, but the effect is milder than in the open-source Vista4D model.
The GEN3C-style variant shows noticeable degradation in visual quality in
some scenes.

\section{User study interface}
\label{app:user-study-interface}
\begin{figure}[H]
  \centering
  \includegraphics[width=\linewidth]{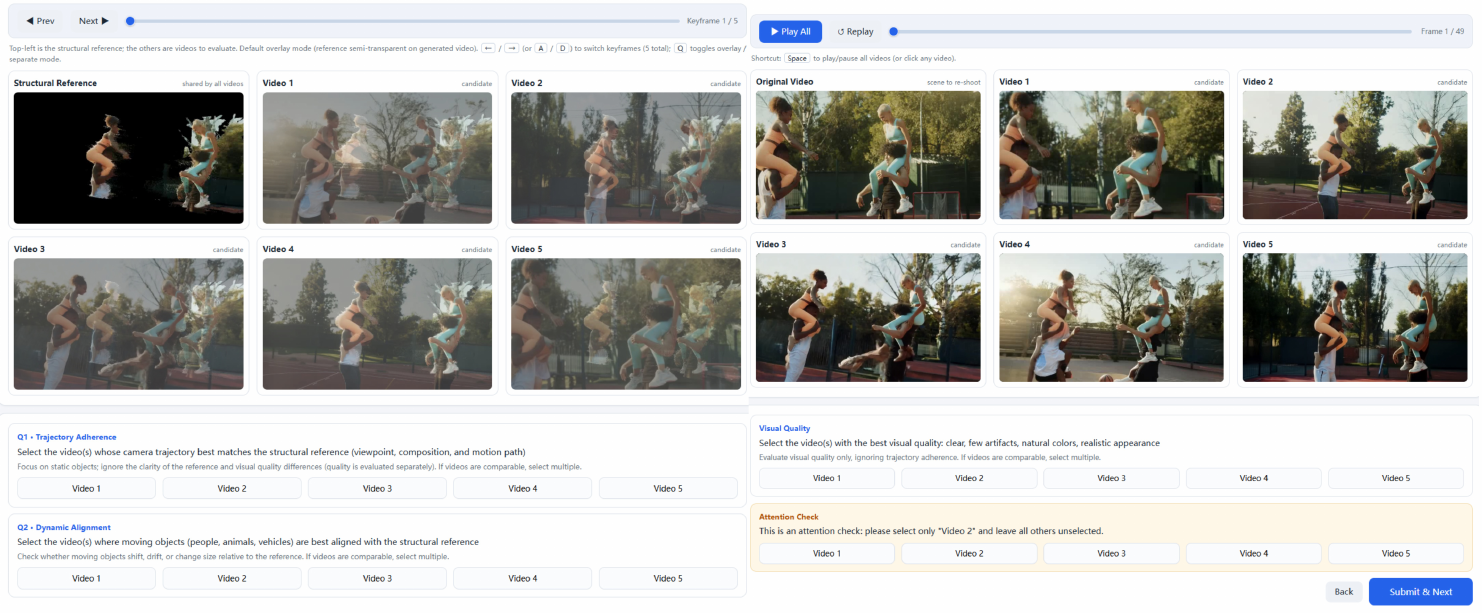}
  \caption{\textbf{User study evaluation interface.} \textbf{Left:} the point
  cloud render is overlaid on each generated video (toggleable with the Q key)
  for the trajectory-following and 3D-consistency questions.
  \textbf{Right:} the full generated video is played without overlay for the
  visual-quality question, presented separately to mitigate bias from the
  preceding overlay-based questions.}
  \label{fig:user_study_interface}
\end{figure}

Fig.~\ref{fig:user_study_interface} shows the user study interface.
The left page evaluates trajectory adherence and 3D consistency of the dynamic
subject; the right page evaluates visual quality.
To assist evaluation, users can toggle an overlay of the point cloud render
on the generated video.
Five keyframes are provided for closer inspection.
Visual quality is assessed on a separate page with the full video playback.
This page is split from the trajectory and consistency evaluation to avoid
bias.
In our pilot study, users tended to pick the same video for all three
criteria.
Separating the pages substantially reduced this effect.

\end{document}